\pdfoutput=1

\documentclass[11pt]{article}

\usepackage{ACL2023}

\usepackage{times}
\usepackage{latexsym}

\usepackage[T1]{fontenc}

\usepackage[utf8]{inputenc}

\usepackage{microtype}

\usepackage{inconsolata}

\usepackage{booktabs} 
\usepackage{graphicx} 
\usepackage{pgffor} 
\usepackage{subcaption} 
\usepackage{xspace} 
\usepackage{multirow}

\newcommand{\arxiv}{\textsc{arXiv}\xspace}

\newcommand{\bump}{\textsc{BUMP}\xspace}
\newcommand{\cnndaily}{\textsc{CNN/DM}\xspace}
\newcommand{\dialsummeval}{\textsc{DialSummEval}\xspace}
\newcommand{\diasumfact}{\textsc{DiaSumFact}\xspace}
\newcommand{\diversumm}{\textsc{DiverSumm}\xspace}
\newcommand{\fib}{\textsc{FIB}\xspace}
\newcommand{\ffci}{\textsc{FFCI}\xspace}
\newcommand{\frank}{\textsc{FRANK}\xspace}
\newcommand{\gofigure}{\textsc{Go Figure}\xspace}

\newcommand{\mlsum}{\textsc{MLSum}\xspace}
\newcommand{\mtsdialog}{\textsc{MTSDialog}\xspace}
\newcommand{\multilexsum}{\textsc{Multi-LexSum}\xspace}
\newcommand{\qmsum}{\textsc{QMSum}\xspace}

\newcommand{\realsumm}{\textsc{REALSumm}\xspace}
\newcommand{\rose}{\textsc{RoSE}\xspace}
\newcommand{\samsum}{\textsc{SAMSum}\xspace}
\newcommand{\seahorse}{\textsc{SEAHORSE}\xspace}

\newcommand{\summedits}{\textsc{SummEdits}\xspace}
\newcommand{\summeval}{\textsc{SummEval}\xspace}

\newcommand{\xlsum}{\textsc{XL-Sum}\xspace}
\newcommand{\xsum}{\textsc{XSum}\xspace}

\title{A Critical Look at Meta-evaluating Summarisation Evaluation Metrics}

\author{Xiang Dai \and Sarvnaz Karimi \and Biaoyan Fang \\
         CSIRO Data61 \\ Sydney, Australia \\ \{dai.dai,sarvnaz.karimi,byron.fang\}@csiro.au}

\begin{document}
\maketitle

\begin{abstract}
Effective summarisation evaluation metrics enable researchers and practitioners to compare different summarisation systems efficiently. Estimating the effectiveness of an automatic evaluation metric, termed \emph{meta-evaluation}, is a critically important research question. In this position paper, we review recent meta-evaluation practices for summarisation evaluation metrics and find that (1) evaluation metrics are primarily meta-evaluated on datasets consisting of examples from news summarisation datasets, and (2) there has been a noticeable shift in research focus towards evaluating the faithfulness of generated summaries. We argue that the time is ripe to build more diverse benchmarks that enable the development of more robust evaluation metrics and analyze the generalization ability of existing evaluation metrics. In addition, we call for research focusing on user-centric quality dimensions that consider the generated summary's communicative goal and the role of summarisation in the workflow. 
\end{abstract}

\section{Introduction}

The evaluation of natural language processing systems is crucial to ensure their effectiveness and reliability in real-world applications. 
It helps compare systems, validate whether the designed properties work as intended, understand the strengths and weaknesses of the underlying model, and often guide iterative improvements~\citep{ribeiro-wu-2020-acl-checklist}. Although human evaluation, especially for natural language generation systems, is considered the most reliable evaluation method~\citep{huang-westlake-2020-emnlp-polytope,iskender-tuberlin-2021-humeval-reliability,khashabi-allenai-2022-emnlp-genie}, automatic evaluation metrics are more widely used due to their cost-effectiveness, ease of use, repeatability, and speed~\citep{graham-2015-emnlp-evaluate-summarization,gehrmann-google-2023-jair-evaluation-practices}. 

In addition to assessing the performance of summarisation systems, automatic summarisation evaluation metrics are also used for other purposes during summarisation system development, such as filtering noisy datasets to improve the quality of training data~\citep{chaudhury-ibm-2022-emnlp-xfactor,aharoni-google-2023-acl-mface}, ranking sampled candidates to output the best summary~\citep{falke-darmstadt-2019-acl-ranking-summaries,chaudhury-ibm-2022-emnlp-xfactor}, and, integrating with reinforcement learning framework as a reward function~\citep{zhang-stanford-2020-acl-correctness-summary,stiennon-openai-2020-neurips-summarize-feedback}. 

A critically important question is \emph{how effective these automatic summarisation evaluation metrics are}. In other words, do the evaluation results obtained using these automatic metrics reflect the genuine quality of the summaries and summarisation systems under examination? For example, \citet{goyal-utexas-2022-zeroshot-news-annotation} conclude that existing automatic metrics cannot reliably evaluate summaries generated using instruct-tuned GPT-3 model~\citep{ouyang-openai-2022-neurips-instruct-gpt}, because they find that GTP-3 summaries receive much lower scores than state-of-the-art fine-tuned models~\citep{liu-liu-2022-acl-brio} on automatic metrics while outperforming them on human evaluation using A/B testing. 

Meta-evaluating summarisation evaluation metrics, especially building resources that enable assessing the automatic metrics, has become urgent and attracted significant research interest~\citep{fabbri-salesforce-2020-tacl-summeval,bhandari-cmu-2020-emnlp-realsumm,clark-google-2023-emnlp-seahorse,liu-fabbri-2023-acl-rose,laban-salesforce-2023-emnlp-summedits}. However, these resources were built and used in various ways, leading to inconsistent and confusing conclusions about the usefulness of these metrics. 

In this position paper, we take a critical look at the practices of meta-evaluating summarisation evaluation metrics. Our paper is organised as follows: we first review recent meta-evaluation practices for summarisation evaluation metrics (Section~\ref{section_preliminaries}); then, in Section~\ref{section_discussion}, we discuss research trends and gaps around four critical decisions that must be made when we assess the automatic metrics, namely, \emph{choosing data to annotate}, \emph{defining quality dimensions}, \emph{collecting human judgements}, and \emph{comparing automatic metrics against human judgements}. Finally, we provide recommendations in Section~\ref{section_summary}. 

\section{Preliminaries}
\label{section_preliminaries}

The task of summarisation aims to generate a summary $\hat{y}$ given a source text $x$, where $\hat{y}$ encapsulates the key information in $x$. The summarisation evaluation metric typically takes the generated summary $\hat{y}$, optionally the source $x$ or a (few) reference summary $y$,  as input and produces a numeric value, which is a proxy of the overall quality or a particular dimension of quality, of $\hat{y}$. 

\subsection{Summarisation Evaluation Metrics}

Summarisation evaluation metrics can be roughly grouped into categories based on what input data they use (e.g., source text, reference summary), what intermediate data they generate (e.g., auto-generated questions based on the source text), what underlying models they rely on (e.g., textual entailment models):
\begin{description}
\item[Summary-only] metrics take the generated summary $\hat{y}$ as input and focus on how well the generated text can be read, e.g., free of syntactic errors or spelling errors~\citep{mani-2001-summeval,goldsack-luo-2023-bionlp-biolaysumm}; 
\item[Similarity-based] metrics take $\hat{y}$ and one or a few reference summaries $y$ as input and measure how similar $\hat{y}$ and $y$ are~\citep{lin-2004-rouge,zhang-cornell-2020-iclr-bertscore}; 
\item[Entailment-based] metrics take both $\hat{y}$ and the source $x$ as input and use entailment models to determine whether the information in $\hat{y}$ is supported by $x$~\citep{laban-schnabel-2022-tacl-summac,honovich-google-2022-naacl-true}; 
\item[QA-based] metrics use both $\hat{y}$ and $x$ and aim to compare the factual information in $\hat{y}$ and $x$ by eliciting answers from them for the same question~\citep{durmus-he-2020-acl-feqa,deutsch-upenn-2021-tacl-qa-metric}; \item[Learning-based] metrics aim to train an evaluation model, using human annotations~\citep{aharoni-google-2023-acl-mface} or weak supervision signals~\citep{kryscinski-salesforce-2020-emnlp-factual-summarization,wu-baidu-2023-acl-wecheck}, that directly outputs the quality score of $\hat{y}$ given $x$; and,
\item[LLM-based] metrics directly instruct large language models to generate the quality score of $\hat{y}$~\citep{tam-unc-2023-acl-fib,shen-alibaba-2023-emnlp-llm-summeval}. 
\end{description}

\subsection{Meta-evaluation of Automatic Metrics}

Estimating the effectiveness and reliability of an automatic evaluation metric is a critically important research question. To distinguish it from summarisation \emph{evaluation}, researchers usually use the term \emph{meta-evaluation} to refer to this task, which is the focus of our position paper. 

Early studies of summarisation meta-evaluation focus on assessing evaluation metrics according to their ability to distinguish between human-written and system-generate summaries~\citep{rankel-maryland-2011-emnlp-ranking-summarization}. However, more recently, a widely accepted belief about meta-evaluation is that an effective evaluation metric should mirror human judgements~\citep{graham-2015-emnlp-evaluate-summarization,huang-westlake-2020-emnlp-polytope,fabbri-salesforce-2020-tacl-summeval,gao-wan-2022-naacl-dialsummeval}. This is often approximated by calculating the correlation between the evaluation results using the automatic evaluation metric $X$ and human judgements $Z$ across a set of summaries generated using various systems. 

Assuming there are $N$ source texts, and $J$ summarisation systems are employed, resulting in a total of $N \times J$ output summaries. We use $d_i$ to represent the $i$-th source text and $s_i^j$ the summary generated by the $j$-th summarisation system on $d_i$. We use $x_i^j$ to represent the score assigned to $s_i^j$ by the evaluation metric $X$ and $z_i^j$ the corresponding human judgement. To measure the correlation between $X$ and $Z$, a correlation function ($\mathbf{Corr}$, such as Pearson, Kendall, or Spearman) is needed. 

\begin{table*}[tb]
    \small
    \centering
    \begin{tabular}{p{0.2\linewidth} p{0.3\linewidth} p{0.2\linewidth} p{0.2\linewidth}}
    \toprule
    & \bf Data & \bf Quality dimensions & \bf Comparison protocol \\
    \midrule
    \summeval~\citep{fabbri-salesforce-2020-tacl-summeval} & \multirow{11}{4.8cm}{Model-generated and (transformed) reference summaries on news articles, such as those in \cnndaily~\citep{nallapati-ibm-2016-conll-summarization}, \xsum~\citep{narayan-edinburgh-2018-emnlp-xsum}, \xlsum~\citep{hasan-buet-2021-acl-xlsum}, and \mlsum~\citep{scialom-dray-2020-emnlp-mlsum}} & Coherence, Faithfulness, Fluency, Relevance & Correlation \\ 
    \\[0.2ex]
    \realsumm~\citep{bhandari-cmu-2020-emnlp-realsumm} & & Relevance & Correlation \\ 
    \\[0.2ex]
    \frank~\citep{pagnoni-cmu-2021-naacl-frank} & & Faithfulness & Correlation \\ 
    \\[0.2ex]
    \ffci~\citep{koto-melbourne-2022-jair-ffci} & & Focus, Coverage, Coherence & Correlation \\ 
    \\[0.2ex]
    \fib~\citep{tam-unc-2023-acl-fib} & & Factual consistency & Ranking  \\ 
    \\[0.2ex]    
    \bump~\citep{ma-dataminr-2023-acl-bump} & & Faithfulness & Ranking, Classification \\ 
    \\[0.2ex]   
    \seahorse~\citep{clark-google-2023-emnlp-seahorse} & & Comprehensibility, Repetition, Grammar, Attribution, Main ideas, and Conciseness & Correlation, Classification \\ \hline 
    \dialsummeval~\citep{gao-wan-2022-naacl-dialsummeval} & \multirow{5}{4.8cm}{Model-generated summaries on dialogues, such as those in \samsum~\citep{gliwa-samsung-2019-samsum}, \qmsum~\citep{zhong-yin-2021-naacl-qmsum}, and \mtsdialog~\citep{abacha-microsoft-2023-eacl-clinical-note}} & Coherence, Consistency, Fluency, Relevance & Correlation \\  
    \\[0.2ex]
    \diasumfact~\citep{zhu-melbourne-2023-acl-diasumfact} & & Factual consistency & Classification \\ 
    \\[0.2ex]
    \citep{abacha-microsoft-2023-acl-note-generation} & & Factual consistency & Correlation \\ \hline
    \gofigure~\citep{gabriel-celikyilmaz-2021-acl-go-figure}  & \multirow{3}{4.8cm}{Model-generated and (transformed) reference summaries on news articles and dialogues} & Faithfulness & Correlation \\ 
    \\[0.4ex]   
    \rose~\citep{liu-fabbri-2023-acl-rose} & & Salience & Correlation \\ \hline 
    \summedits~\citep{laban-salesforce-2023-emnlp-summedits} & \multirow{3}{4.8cm}{Model-generated summaries, and LLM-edited reference summaries on diverse domains, such as news articles, scholarly articles, meeting transcripts, government reports, legal bills, etc.} & Faithfulness & Classification \\ 
    \\[1.4ex]   
    \diversumm~\citep{zhang-ed-2024-eacl-infuse} & & Faithfulness & Classification \\ 
    \\[1.4ex]   
    \citep{ramprasad-krishna-2024-factuality-varied-domains} & & Faithfulness & Correlation \\ 
    \bottomrule
    \end{tabular}
    \caption{A summary of recent benchmarks for meta-evaluating summarisation evaluation metrics.}
    \label{table_existing_benchmarks}
\end{table*}

\paragraph{System-level} protocol aggregates the evaluation scores for a given summarisation system first via:
\begin{equation}
    x^j = \frac{1}{N} \sum_{i=1}^N x_i^j,
\end{equation}
where $x^j$ is an approximation of the judgement of the $j$-th summarisation system by metric $X$.
Similarly, the human judgement can be aggregated via:
\begin{equation}
    z^j = \frac{1}{N} \sum_{i=1}^N z_i^j.
\end{equation}
Then, the two lists of judgements, each containing $J$ values, are taken as input to calculate the system-level correlation coefficient and the corresponding $p$-value:
\begin{equation}
    r, p = \mathbf{Corr} \biggl( \biggr[ x^1, \cdot, x^J \biggr], \biggr[ z^1, \cdot, z^J \biggr] \biggl).
\end{equation}
\paragraph{Summary-level} protocol calculates the correlation between $X$ and $Z$ on each summary first:
\begin{equation}
    r_i, p = \mathbf{Corr} \biggl( \biggr[ x_i^1, \cdot, x_i^J \biggr], \biggr[ z_i^1, \cdot, z_i^J \biggr] \biggl),
\end{equation}
and then apply an average operation to obtain the summary-level correlation coefficient:
\begin{equation}
    r = \frac{1}{N} \sum_{i=1}^N r_i.
\end{equation}

In addition to this common ``correlation'' perspective, recent studies focusing on evaluating the faithfulness of summaries also use classification or ranking protocols. That is, the generated summary (or more fine-grained elements, such as a sentence) is labelled by human annotators, for example as ``faithful'' or ``unfaithful'', and then automatic evaluation metrics are evaluated by whether they can predict accurately the label of a given summary (i.e., classification) or assigning a higher score to the faithful summary than the unfaithful summary (i.e., ranking). 

We summarise recent benchmarks for meta-evaluating summarisation evaluation metrics in Table~\ref{table_existing_benchmarks}. A more detailed description of these benchmarks can be found in the Appendix~\ref{section_existing_meta_evaluation_datasets}. 

\section{Discussion}
\label{section_discussion}
We identify there are four critical decisions that must be made when we assess the automatic metric: (1) what source texts and summaries to use; (2) what quality dimensions to consider; (3) how to collect human judgements; and, (4) how to compare the automatic metric against human judgements. In this section, we discuss research trends and gaps around these four aspects. 

\subsection{Choosing Data to Annotate}
\begin{figure}[tb]
    \centering
    \includegraphics[width=\linewidth]{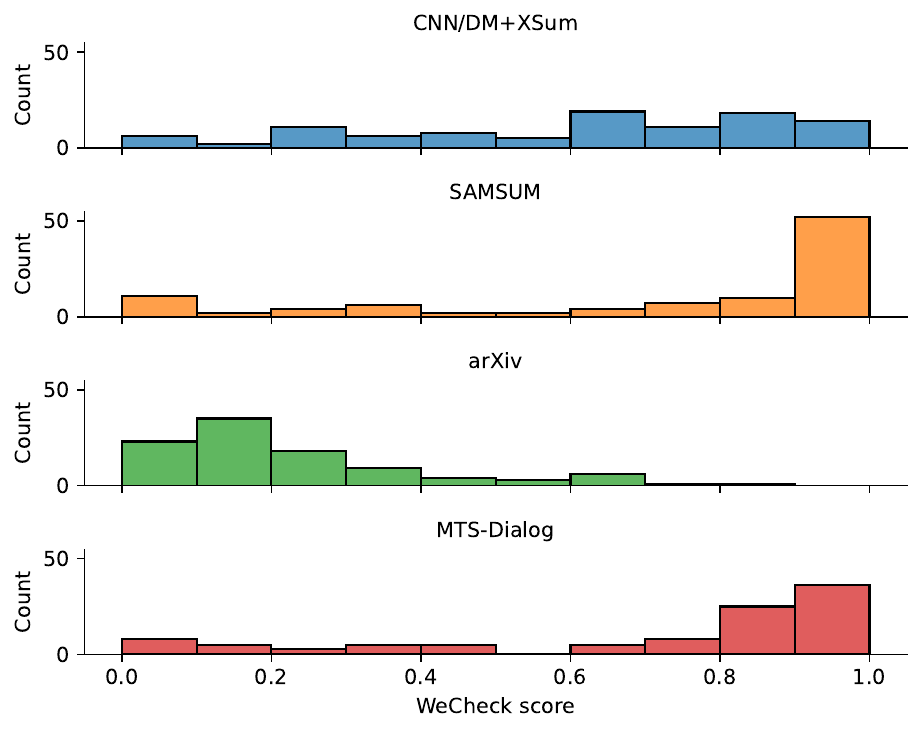}
    \caption{The distribution of consistency scores, measured using WeCheck~\citep{wu-baidu-2023-acl-wecheck}, between source text and reference summary from different datasets. A score of $1$ indicates a higher consistency level, while $0$ indicates inconsistency. \cnndaily and \xsum datasets~\citep{zhang-ladhak-2024-tacl-benchmark-llm-summarization} include news articles, \samsum~\citep{gliwa-samsung-2019-samsum} messenger-like conversations, \arxiv~\citep{cohan-dernoncourt-2018-naacl-long-summarization} scholarly articles, and \mtsdialog~\citep{abacha-microsoft-2023-eacl-clinical-note} from Doctor-Patient encounters. }
    \label{figure_wecheck_various_datasets}
\end{figure}

\paragraph{Source texts}
From Table~\ref{table_existing_benchmarks}, we can see that most of the widely used meta-evaluation benchmarks use source texts from news summarisation datasets, followed by dialogue summarisation datasets. 
This is not ideal because, first, evaluation metrics tailored for evaluating news articles and summaries may not be portable to other domains due to the lack of respective resources. For example, QA-based evaluation metrics~\citep{wang-nyu-2020-acl-factual-consistency,durmus-he-2020-acl-feqa} usually start from extracting named entities  (e.g., person names) from source text and/or generated summary, around which questions are generated. However, entities of interest vary in different domains, and effective named entity recognition tools may not exist for specialised entity categories in niche domains, making these evaluation metrics hard to use. Secondly, the distribution of automatic evaluation scores usually differs across texts from various domains (Figure~\ref{figure_wecheck_various_datasets}), and the generalisation ability of these evaluators, which are calibrated to the news domain, is underexplored~\citep{laban-salesforce-2023-emnlp-summedits}. Finally, evaluation metrics usually show different correlation trends in different datasets, making their practical utility unclear. For example, \citet{ramprasad-krishna-2024-factuality-varied-domains} find that both QA-based and NLI-based evaluation metrics correlate well (Spearman's rank correlation coefficients of $0.45 \sim 0.59$) with human judgements on examples from news domain, but no correlation on biomedical domain (coefficients of $-0.03 \sim 0.11$). 


\begin{figure}[tb]
    \includegraphics[width=1.1\linewidth]{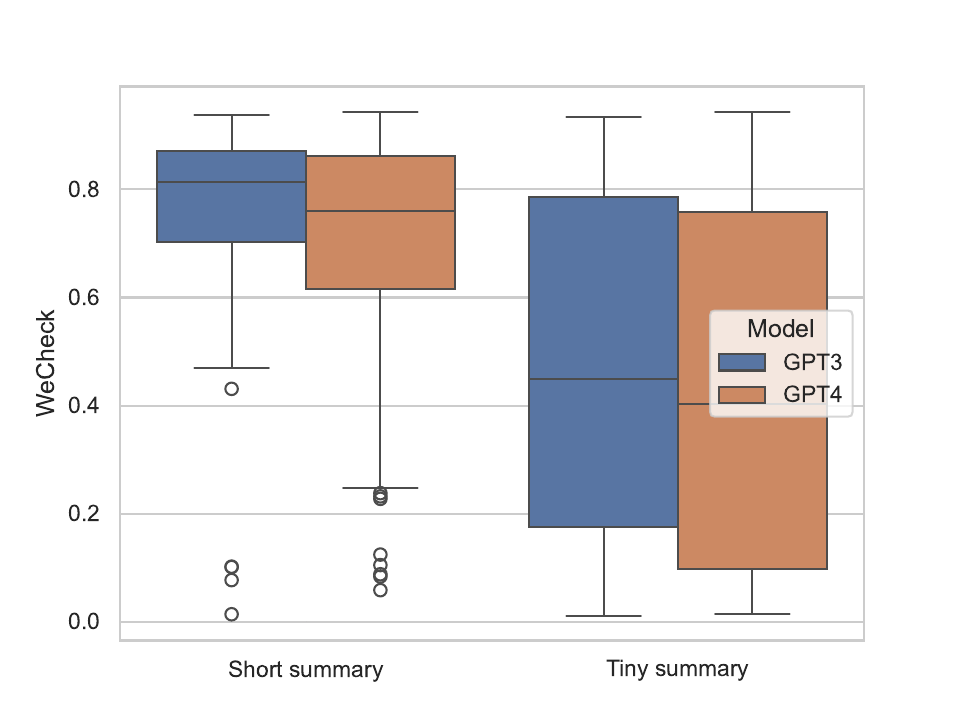}
    \caption{Evaluation results using WeCheck~\citep{wu-baidu-2023-acl-wecheck} on two tasks proposed in Multi-LexSum~\citep{shen-allenai-2022-neurips-multi-lexsum}, where summaries are generated at different target levels of granularity: tiny (25 words, on average), and short (130 words). Prompts used to generate summaries can be found in Appendix Section~\ref{section_implementation_details}.}
    \label{figure_multi_lexsum_two_tasks}
\end{figure}

\paragraph{Output summaries}
A common strategy for collecting summaries is assembling outputs from diverse summarisation systems, which are expected to cover different error types. For example, \citet{clark-google-2023-emnlp-seahorse} collect summaries from models of various sizes (e.g., 220M parameters of T5~\citep{raffel-google-2020-jmlr-t5} and 540B parameters of PaLM~\citep{aakanksha-google-2024-jmlr-palm}) and employ both $1$-shot in-context learning and fine-tuning approaches to generate the summaries. They also select both fully optimised and under-trained (i.e., trained for only 250 steps) checkpoints, ensuring differences in model quality. 

Although these studies seek to diversify the summarisation systems, they often operate under a uniform summarisation formulation. In other words, the communicative goal and user preferences (e.g., the desired style and summary length) are disregarded when generating the summary. For example, \citet{ramprasad-krishna-2024-factuality-varied-domains} use the same prompt `Article: [article]. Summarize the above article:' for generating summaries across domains. This simplified task formulation might be problematic when translating findings to build real-world summarisation applications. Summarisation involves compressing information in the source text by definition, and one key factor in this process is the compression ratio. Figure~\ref{figure_multi_lexsum_two_tasks} shows that, under various constraints such as summary length~\citep{koh-ju-2022-emnlp-long-summarization}, evaluation metrics may exhibit varying characteristics since generating shorter summaries (with a higher compression ratio) and evaluating these summaries is more challenging. 


\paragraph{Summary}
Because of the lack of meta-evaluation benchmarks covering various data distributions (i.e., source texts from different domains and output summaries from different systems under different task constraints), NLP practitioners may take the risk of overestimating the generalisation ability of automatic metrics~\citep{chen-fudan-2021-emnlp-factuality}. That is, practitioners may employ the top-performing evaluation metrics, e.g., for evaluating news summaries, and hope they work well for evaluating other types of summaries. To fill this gap, we call for building more diverse benchmarks that enable building more robust evaluation metrics and analysing the generalisation ability of existing evaluation metrics across different domains. 

\subsection{Defining Quality Dimensions}

\citet{mani-2001-summeval} divide the summarisation evaluation into two categories: intrinsic evaluation---testing the summarisation system as of itself---and extrinsic evaluation---testing based on how the generated summary affects the completion of some downstream tasks (e.g., efforts required to post-edit the generated summary to an acceptable, task-dependent state). We notice that most---if not all---recent benchmarks focus on quality dimension relating to the intrinsic evaluation but overlook the extrinsic evaluation. 

From Table~\ref{table_existing_benchmarks}, we observe quality dimensions considered in recent benchmarks can be roughly grouped into two categories: (1) content quality, concerning to which extent the generated summary \emph{accurately} reflects the most \emph{important} information in the source text, and (2) language quality (e.g., coherency, fluency, comprehensibility) of the generated summary itself. We also notice that there is a shift in research focus towards content quality, especially the faithfulness of generated summaries. 

\citet{fonseca-cohen-2024-scientific-summarizers} argue that summarisation evaluation should consider the variability in communicative intentions. They choose three intentional aspects: conciseness (e.g., \textit{Write a summary of the article above in 3 sentences}), narrative perspective (e.g., \textit{Write in third person}), and keyword coverage (e.g., \textit{Focus on the keywords: Thompson, sampling, sequential, variational}). They  define \emph{intention control metrics} to assess whether the generated summaries follow these intentions accurately. \citet{zhang-ladhak-2024-tacl-benchmark-llm-summarization} also point out that the summarisation evaluation should depend on the application scenarios and align with user values. They argue that, for example, the bullet point style summaries in \cnndaily~\citep{nallapati-ibm-2016-conll-summarization} are rated by human annotators with low coherence scores; however, they may suffice for being displayed on news websites. 

It is also worth noting that some quality dimensions are user-centric by nature, but most existing studies have overlooked the subjectivity of these dimensions. For example, when we define the readability of plain-language summaries of scientific articles \citep{goldsack-sheffield-2022-emnlp-summarisation-science}, the end users' language and domain background should be taken into consideration. Another example is clinical conversation summarisation \citep{abacha-microsoft-2023-acl-note-generation}; depending on whether the summary is provided for the clinicians or the patients, the same quality dimension (e.g, comprehensibility) should be defined differently. 


\paragraph{Subtle differences behind the same term}

We observe a clear shift of research focus towards the content quality, especially faithfulness, of summarisation, mainly because recent LLM-based summarisation models have shown a remarkable capability to produce text of high language quality but still struggle with generating accurate content in a conditional-generation setting~\cite{gao-wan-2022-naacl-dialsummeval}. 

However, we also notice that different studies may investigate the same quality dimension following slightly different definitions, resulting in confusing conclusions. For example, \citet{fabbri-salesforce-2020-tacl-summeval} define ``consistency'' as ``whether the facts in the summary are consistent with the facts in the original article'' but also ``consider whether the summary does reproduce \emph{all} facts accurately and does not makeup \emph{unture} information''. \citet{honovich-google-2022-naacl-true} define a text to be factually consistent to its grounding text (i.e., source text) ``if all the factual information it conveys is consistent with the factual information conveyed by the grounding text'' but ``exclude personal and social statements''. These subtle differences usually result in different judgements on the same summary due to ``partial faithful'' or ``factual but not faithful'' issues. 


\paragraph{Summary}
We believe quality dimensions considered in the recent benchmarks are too narrow to reflect the various application scenarios where summarisation is used. Even worse, there is no census on the precise definition of these quality dimensions---different terms reflect the same underlying meaning and the same term refers to slightly different meanings---making the comparison against previous work difficult and unreliable. 

\subsection{Collecting Human Judgements\label{section_collect_annotations}}


\paragraph{Who are expert annotators?}

Most previous studies, especially those that focus on news summarisation, refer to expert annotators as people who have experience in summarisation or NLP.
Correspondingly, annotation guidelines are also heavily linguistic-oriented, for instance in their error categories and examples. 
For example, \citet{pagnoni-cmu-2021-naacl-frank} collect human annotations based on a typology of factual errors, including ``Relation Error'', ``Entity Error'', ``Circumstance Error'', ``Discourse Link Error'', etc.
Although this perspective can help developers understand the weaknesses of different summarisation models by examining the common errors these systems may generate, we argue that these errors may not always reflect real users' perspectives; instead, the real writers and readers of summaries should be more involved in the annotation process and the development of annotation guidelines.


\paragraph{Trade-off between annotation quality and cost}


Crowdsourcing is a common approach to reduce the time and cost associated with data annotation, though it often comes at the cost of sacrificing the reliability of the collected annotations. Most recent efforts that build meta-evaluation benchmarks rely on crowd annotators because crowd annotations can typically be collected quickly. In contrast, expert annotators may require significantly more time, even when fully dedicated to the annotation task. For example, \citet{gao-wan-2022-naacl-dialsummeval} reported that they initially conducted the annotation via a crowd-sourcing platform and collected $7,000$ annotations from five different annotators in one day. In contrast, it took approximately $10$ days to collect $4,200$ annotations from three student annotators. 




Using LLMs as surrogate evaluators or combining LLM-as-evaluators with human evaluation to obtain an unbiased estimator with a lower cost than human evaluation alone is a promising but controversial research direction. Its effectiveness needs careful investigation as it depends not only on the correlation between the human and LLM-as-evaluator judgements but also on the choice of evaluation prompts (e.g., reference-free evaluation, pair-wise comparison, Likert survey)~\citep{chaganty-stanford-2018-acl-debiasing-metrics}. \citet{deutsch-upenn-2022-emnlp-reference-free} point out that when we use one generation model to evaluate another, they are biased against higher-quality outputs, including those written by humans. \citet{liu-sheffield-2023-llm-evaluators,panickssery-feng-2024-llm-evaluators} also show that LLM-as-evaluators may have the problem of self-preference---they favour their own outputs or outputs from similar model families.

Ensuring annotation quality and detecting noisy annotations are then essential to building a reliable benchmark using crowd-sourcing or combining LLM-as-evaluators with human evaluation. However, we notice that only a limited number of quality control practices were commonly adopted in eliciting human annotations, such as filtering annotators based on their previous experience~\citep{liu-fabbri-2023-acl-rose}, providing annotator training~\citep{aharoni-google-2023-acl-mface} and measuring inter-annotator agreement~\citep{laban-salesforce-2023-emnlp-summedits}. Moreover, many studies overlook this issue and place blind trust in the collected data. For instance, \citet{koto-melbourne-2022-jair-ffci} found that only $7$ out of $71$ papers on summarisation human evaluation describe quality control mechanisms used. 

Another overlooked practice is reporting failed attempts, which we believe can provide valuable insights to the following studies. For example, \citet{gao-wan-2022-naacl-dialsummeval} hired $5$ annotators using a crowdsourcing platform to assess summaries generated from $14$ different summarisation models on a Likert scale from $1$ to $5$. However, the model scores, which are calculated by averaging across $5$ annotators on $100$ summaries, are very close to each other (e.g., the averaged consistency score of the worst model is $3.206$, whereas the best is $3.400$), which they believe does not reflect reality. 


%



\paragraph{The role of reference summary}
Intuitively, some quality dimensions can be assessed by reading the summary only. For example, \citet{goldsack-sheffield-2022-emnlp-summarisation-science} instruct the annotators to rate \emph{layness} (to what extent is the summary comprehensible to a non-expert) using a $1$ to $5$ Likert scale. However, these annotations usually suffer from inconsistency issues, as even the same annotator may make different assessments at different times. 

Relative assessment, instead of direct assessment, is generally considered to improve agreement among annotators~\citep{novikova-heriot-watt-2018-naacl-rankme}. However, existing work uses reference summaries to aid human judgements mainly from a cost-saving consideration because annotators can rate the quality of a generated summary by comparing it against a short reference summary without reading a relatively long source text. Also, using a reference summary may reduce the annotation complexity for non-expert annotators. For example, \citet{koto-melbourne-2022-jair-ffci} argue that assessing \emph{relevance}---the generated summary concisely captures all salient information---without a reference summary is difficult, as it requires annotators to implicitly construct their own summary of the source text.

However, the impact of reference summaries on human judgements and thus on meta-evaluation results is not well understood and examined.
Regarding the same set of quality dimensions (\emph{fluency}, \emph{coherence}, \emph{faithfulness} and \emph{relevance}), \citet{fabbri-salesforce-2020-tacl-summeval} provide annotators with summaries grouped in sets of $6$ (i.e., $1$ reference summary and $5$ model-generated summaries), where the reference summary plays as an anchor between groups. But \citet{zhuang-adelaide-2024-naacl-evaluation} find that annotators tend to assign a lower score to the summary if it is shown along with a reference summary---even with a false reference summary.
Automatic metric performance might also differ greatly depending on whether reference summaries are used during human annotations.
For example, \citet{liu-fabbri-2023-acl-rose} find that reference-based metrics generally perform better when they are compared against human judgements collected using protocols with reference summary but can have negative correlations with those without using reference summary.

\paragraph{Human preferences vs quality judgement}
Instead of scoring summaries based on the description of quality dimensions, \citet{goyal-utexas-2022-zeroshot-news-annotation} adopt the approach of soliciting human preferences among summaries. However, this approach may be questionable when involving summaries generated using LLMs, which are usually pre-trained with human preference feedback. \citet{liu-fabbri-2023-acl-rose} point out that LLMs may have learned the prior preferences of human annotators but not necessarily captured the task-specific quality of summaries. \citeauthor{liu-fabbri-2023-acl-rose} designed two studies, asking human annotators: (a) to evaluate the summary without knowing the input text and (b) to evaluate if the summary covers the salient information of the input text. Results show that LLM-generated summaries received higher scores than human-written summaries under the first study, and the scores obtained from the first study are a good predictor of the results of the second study (Pearson's correlation of $0.926$ between these two results). \citet{zhang-ladhak-2024-tacl-benchmark-llm-summarization,shaib-adobe-2024-summarization-annotation} also identify that annotators usually have their own consistent preference (e.g., based on summary length), when simply asked to rank the summaries. 

\paragraph{Summary}
Given the costly nature of eliciting human judgements and the rapid pace of ongoing development in summarisation models, we believe there is an urgent need to standardise human evaluation practices. Developing a mechanism for producing reproducible human judgements over time and across different annotators~\citep{khashabi-allenai-2022-emnlp-genie} is paramount because it allows the collected resources to be reusable and easily extensible to new summarisation models. The resulting resources, which are more comprehensible, enable the development of effective and robust automatic metrics. 

\subsection{Comparing Automatic Metrics Against Human Judgements}

\paragraph{Is a high correlation with human judgements enough to indicate the effectiveness of automatic metrics?}
A common way of reporting the effectiveness of automatic metrics is to tabulate the correlation between the results obtained using automatic metrics and human judgements, and metrics that achieve higher correlation are considered to be better~\citep{fabbri-salesforce-2020-tacl-summeval,ramprasad-krishna-2024-factuality-varied-domains}. However, \citet{ernst-amazon-2023-emnlp-multiple-quality} find that some evaluation metrics, although highly correlating with human judgements on a particular quality dimension, are, in fact, ineffective in measuring the considered dimension.
For example, reference-based evaluation metrics correlate well with human judgements on \emph{Fluency} and \emph{Consistency} in the \summeval~\citep{fabbri-salesforce-2020-tacl-summeval} benchmark but fail to detect even drastic summary corruptions, such as replacing all verbs with lemma form (resulting in ungrammatical summaries) and all person names with different names from the source text (resulting in unfaithful summaries). 

One reason behind this phenomenon is that human judgements across quality dimensions may correlate with each other (Table~\ref{table_summeval_agreement} in Appendix~\ref{section_appendix_additional_results}). Therefore, it is necessary to rule out the impact of confounding factors when comparing automatic metrics against human judgements for a particular quality dimension. \citet{ernst-amazon-2023-emnlp-multiple-quality} propose a bucketing-based approach where they divide all document-summary pairs into buckets where the human judgements of an anchor dimension have low variance; the correlations are calculated inside each bucket and then averaged with weights according to bucket size, resulting in more reliable meta-evaluation results for dimensions other than the anchor dimension. 

Another reason is that most existing benchmarks include summaries generated from systems of varying quality. Therefore, high correlation is usually attributed to the capability of distinguishing between systems with large performance gaps. \citet{deutsch-upenn-2022-naacl-system-level-correlations,liu-fabbri-2023-acl-rose,shen-alibaba-2023-emnlp-llm-summeval} point out that discriminating between systems of similar quality is much more difficult than between systems of diverse quality, and a good metric should reliably indicate a difference in quality when a small difference in evaluation scores is observed. For example, \citeauthor{deutsch-upenn-2022-naacl-system-level-correlations} found that the average improvement over baseline models reported in recent papers on the \cnndaily~\citep{nallapati-ibm-2016-conll-summarization} dataset was ROUGE-1 score of $0.5$. However, the correlation of ROUGE-1 to human judgements is near $0$ when ranking systems whose evaluation scores are so close.  On the other hand, a large gap (e.g., $5$-$10$) of ROUGE scores does correctly rank system pairs, enabling ROUGE to achieve moderately strong correlations on standard benchmarks. 

\paragraph{Statistical Power} 
concerns the chance a \emph{significant} difference (e.g., evaluation metrics score differently in a meta-evaluation benchmark) will be observed, given there is a real difference (i.e., genuinely different evaluation metrics)~\citep{card-stanford-2020-emnlp-power-analysis}. \citet{deutsch-upenn-2021-tacl-stat-analysis-summeval} find that high uncertainty (large confidence intervals) exists when evaluating automatic metrics using existing benchmarks. This is also observed in human evaluation of similar-performing systems~\citep{liu-fabbri-2023-acl-rose}.
Although increasing the dataset's sample size (but requiring a significant human effort) can effectively raise statistical power~\citep{shaib-adobe-2024-summarization-annotation}, other cheap alternatives are needed. For example, \citet{deutsch-upenn-2022-naacl-system-level-correlations} propose to calculate automatic scores on a much larger set instead of only the subset of summaries judged by humans. 







\paragraph{Summary}

We argue that assessing the effectiveness of automatic metrics can be conducted in multiple stages, each requiring different levels of human annotation effort. First, evaluation metrics should be tested on their effectiveness in detecting significant errors, e.g., corruptions in human-written summaries~\citep{gabriel-celikyilmaz-2021-acl-go-figure,chen-fudan-2021-emnlp-factuality,ernst-amazon-2023-emnlp-multiple-quality}. Secondly, they can be meta-evaluated against existing human judgements on summaries from systems of varying quality. Thirdly, human judgements should be constantly gathered on summaries generated using state-of-the-art systems, presumably of closing quality~\citep{peyrard-2019-acl-summarization-evaluation} and automatic metrics should be tested on discriminating these systems. Finally, metrics should be tested against reproducing human preferences between pairs of summaries (i.e., summary-level effectiveness) and the capability of identifying more fine-grained problems~\citep{chen-fudan-2021-emnlp-factuality}. 

\section{Related Work}


Similar to summarisation, other natural language generation tasks, such as machine translation (MT), dialogue, and data-to-text generation, also have a long history of employing automatic evaluation metrics, such as BLEU~\citep{papineni-ibm-2002-acl-bleu} and METEOR~\citep{banerjee-lavie-2025-meteor}, to assess the quality of machine-generated texts. Assessing the effectiveness and reliability of these automatic metrics is also an active research area, and regular shared tasks (e.g., WMT Metrics Shared Task\footnote{\url{https://www2.statmt.org/wmt24/metrics-task.html}}) are organised to encourage researchers to explore the strengths and weaknesses of automatic metrics. Unfortunately, similar efforts to meta-evaluate summarisation evaluation metrics were unsustained, partially due to the complexity of the summarisation task itself. As observed by~\citet{graham-2015-emnlp-evaluate-summarization}, although there are obvious parallels between summarisation and machine translation (MT), methodologies applied to meta-evaluate MT metrics have not been well explored in summarisation.

With the advancement of large-scale generative models, evaluating text generated by LLMs and meta-evaluate corresponding evaluation metrics have also attracted significant interests~\citep{gehrmann-adewumi-2021-gem,chen-hkust-2023-neurips-felm,li-ruc-2023-emnlp-halueval,pal-saama-2023-conll-med-halt,mishra-washington-2024-fine-grained-hallucination}.
These studies usually concern similar quality dimensions as those in the summarisation field, and have a similar desire to find cost-effective ways to collect human judgements.

\section{Conclusions and Recommendations}
\label{section_summary}
In this position paper, we critically examine the practices of meta-evaluating summarisation evaluation metrics in the literature. We identify several avenues in the field that can be further improved, regarding: \emph{choosing data to annotate}, \emph{defining quality dimensions}, \emph{collecting human judgements}, and \emph{comparing automatic metrics against human judgements}. 

For practitioners aiming to assess the effectiveness of automatic metrics for their particular use case, we suggest starting by considering the role a summarisation system plays in the real-world workflow. This includes identifying the readers of the generated summaries, understanding what information they seek, and what decisions they might make after reading the summary. Once we have a clear picture of this, we can create document-summary pairs that meet requirements and focus on the quality dimensions that end-users value most. Human judgements can be collected from real end-users regarding both their perceived quality of the summary and the effect of these summaries on the actual downstream tasks they perform. Finally, automatic evaluations can be assessed depending on the purposes of the evaluation, such as determining which summarisation system is better (system-level correlation is informative), choosing the best summary from multiple candidates (summary-level correlation/ranking), and detecting problematic summaries (binary classification).

For researchers who aim to develop meta-evaluation resources and novel evaluation metrics, we believe it is time to build more diverse benchmarks using data sampled from different domains and considering various summarisation constraints. That is to say, the generality of evaluation metrics should be tested to mitigate the risk of overestimating the effectiveness of automatic metrics across domains and applications~\citep{gabriel-celikyilmaz-2021-acl-go-figure}. Secondly, we believe there is an urgent need to standardise human evaluation practices to ensure reproducible human judgements over time and, more importantly, to make the collected resources extensible to new summarisation models. We recommend some best (basic) practices: (1) being aware of previous work and reusing previous resources (taxonomy, guideline, interface, etc.) whenever possible~\citep{tang-utexas-2023-acl-aggrefact}; (2) adopting quality controls, such as training annotators to make sure they understand the annotation task, filtering out unqualified annotators and their annotations, etc~\citep{koto-melbourne-2022-jair-ffci}; (3) documenting the creation process (e.g., preprocessing, annotating) and recommended uses~\citep{gebru-morgenstern-2021-datasheets}. Finally, we argue that claims on the usefulness of evaluation metrics should be made based on comprehensive and reliable assessment under various usage scenarios, such as detecting summaries with significant errors, distinguishing summary systems of closing quality, or identifying more fine-grained issues in the generated summary.

\section*{Limitations}
The main limitation of this study is that we did not contribute any new resources or procedures for meta-evaluation. This study states our opinions based on our educated review of the current literature. 

\bibliography{references-short}
\bibliographystyle{acl_natbib}

\appendix

\section{Implementation Details}
\label{section_implementation_details}

\subsection{Prompts used to generate summaries on tasks proposed in \multilexsum}
\begin{itemize}
    \item Short summary (GPT-3.5)
    
    System: As a junior legal intern, please craft a summary (approximately 130 words) for the given legal case.
    
    User: [article]
    \item Tiny summary (GPT-3.5)
    
    System: As a junior legal intern, please craft a summary (approximately 25 words) for the given legal case.
    
    User: [article]
    \item Short summary (GPT-4)
    
    System: As a senior legal professional, please craft a summary (approximately 130 words) for the given legal case.
    
    User: [article]
    \item Tiny summary (GPT-4)
    
    System: As a senior legal professional, please craft a summary (approximately 25 words) for the given legal case.
    
    User: [article]
\end{itemize}

\section{Additional Results}
\label{section_appendix_additional_results}

Table~\ref{table_summeval_agreement} show the correlation between different quality dimensions within the same annotator group and across different groups for the same dimension.

\begin{table}[tb]
    \centering
    \setlength{\tabcolsep}{3pt}
    \begin{tabular}{r cccc c}
    \toprule
     & Coher. & Faith. & Fluen. & Rele. & Expert \\ 
     \midrule
     & \multicolumn{5}{c}{Expert annotators} \\
     \midrule
Coherence & - & 0.300 & \underline{0.544} & \underline{0.700} & \textbf{\underline{0.877}} \\ 
Faithfulness & 0.300 & - & \underline{0.594} & \underline{0.500} & \textbf{\underline{0.900}} \\ 
Fluency & \underline{0.544} & \underline{0.594} & - & \underline{0.745} & \textbf{\underline{0.810}} \\ 
Relevance & \underline{0.700} & \underline{0.500} & \underline{0.745} & - & \textbf{\underline{0.857}} \\ 
     \midrule
     & \multicolumn{5}{c}{Crowd annotators} \\
     \midrule
Coherence & - & 0.310 & \textbf{\underline{0.500}} & \underline{0.393} & -0.083 \\ 
Faithfulness & 0.310 & - & \textbf{0.343} & 0.168 & 0.059 \\ 
Fluency & \textbf{\underline{0.500}} & 0.343 & - & 0.326 & 0.142 \\ 
Relevance & \textbf{\underline{0.393}} & 0.168 & 0.326 & - & -0.159 \\ 
     \bottomrule
    \end{tabular}
    \caption{System-level Kendall’s $\tau$ correlation coefficients between different quality dimensions within the same annotator group, and correlation coefficients between different annotator groups for the same quality dimension. \underline{underline}: the correlation coefficient is significant ($p \le 0.05$). The human judgements are from \summeval~\citep{fabbri-salesforce-2020-tacl-summeval}.}
    \label{table_summeval_agreement}
\end{table}

\section{Meta-evaluation Benchmarks}
\label{section_existing_meta_evaluation_datasets}

\paragraph{\summeval}
\citet{fabbri-salesforce-2020-tacl-summeval} assembled a collection of summaries generated by $16$ models trained on the \cnndaily~\citep{nallapati-ibm-2016-conll-summarization} dataset and collect human judgements from $3$ expert judges and $5$ crowd-source workers. Judges were asked to evaluate the summaries along four dimensions: relevance (concerning the selection of important content), consistency (concerning factual alignment between the summary and the source), fluency (concerning the quality of individual sentences), and coherence (concerning the collective quality of all sentences). 

\paragraph{\realsumm}
\citet{bhandari-cmu-2020-emnlp-realsumm} released a dataset of human judgements on the relevance of summaries collected from $25$ neural summarization systems. \citeauthor{bhandari-cmu-2020-emnlp-realsumm} create Semantic Content Units (SCUs) for each reference summary and then hire crowd workers to annotate each generated summary, determining whether each SCU can be inferred from the generated summary. 

\paragraph{\frank}
\citet{pagnoni-cmu-2021-naacl-frank} devise a typology of factual errors (e.g., predicate errors, entity errors, circumstance errors, etc.) and use it to collect human annotations of generated summaries for the \cnndaily~\citep{nallapati-ibm-2016-conll-summarization} and \xsum~\citep{narayan-edinburgh-2018-emnlp-xsum} datasets. They conduct the annotation task on the Amazon Mechanical Turk platform and found a nearly perfect agreement (a Cohen Kappa of $0.86$) between the majority class of the three crowd annotators and one expert annotator on $20$ summaries.

\paragraph{\gofigure}
\citet{gabriel-celikyilmaz-2021-acl-go-figure} introduce a meta-evaluation framework for evaluating factuality evaluation metrics. \citeauthor{gabriel-celikyilmaz-2021-acl-go-figure} build one diagnostic dataset that consists of transformed reference summaries with simulated factuality errors (i.e., pronoun entity errors, verb tense or negation errors, intrinsic entity errors, extrinsic entity errors, sentiment errors, false quotes). They also use fine-tuned T5 summarization models to generate summaries and annotate them for fine-grained factual errors based on the above-mentioned error types.


\paragraph{\dialsummeval}
\citet{gao-wan-2022-naacl-dialsummeval} sample $100$ dialogues from the \samsum~\citep{gliwa-samsung-2019-samsum} test set and evaluate the summaries generated by $14$ summarization models. Three college students fluent in English were recruited to assess the relevance, consistency, fluency and coherence quality of generated summaries.

\paragraph{\bump}
\citet{ma-dataminr-2023-acl-bump} introduce a dataset of $889$ summary pairs, where a single error is introduced to a reference summary from the \cnndaily~\citep{nallapati-ibm-2016-conll-summarization} dataset to produce an unfaithful summary.
\citeauthor{ma-dataminr-2023-acl-bump} define a taxonomy of seven unfaithful error types (i.e., intrinsic/extrinsic predicate error, intrinsic/extrinsic entity error, intrinsic/extrinsic circumstance error, and coreference error) and instruct annotators to introduce errors of a specific type.

\paragraph{\rose}
\citet{liu-fabbri-2023-acl-rose} propose a human evaluation protocol for evaluating the salience of summaries that is more objective by dissecting the summaries into fine-grained content units and defining the annotation task based on those units. Using the protocol, \citeauthor{liu-fabbri-2023-acl-rose} curate a large human evaluation dataset consisting of $22,000$ summary-level annotations over $28$ systems on samples from \cnndaily~\citep{nallapati-ibm-2016-conll-summarization}, \xsum~\citep{narayan-edinburgh-2018-emnlp-xsum}, and \samsum~\citep{gliwa-samsung-2019-samsum}.

\paragraph{\seahorse}
\citet{clark-google-2023-emnlp-seahorse} collect annotations along $6$ dimensions: comprehensible (read and understood by the rater), repetition (free of unnecessarily repeated information), grammar (grammatically correct), attribution (fully attributable to the source article), main ideas (captures the main ideas of the source article), and, conciseness (concisely represents the information in the source article). Annotators can answer 'Yes,' 'No,' or 'Unsure' to the first three questions given only the summary, and the last three questions given both the article and the summary. Their annotations provide both a benchmark for meta-evaluation but also a resource for training learning-based evaluation metrics.

\paragraph{\summedits}
\citet{laban-salesforce-2023-emnlp-summedits} propose a new protocol for creating inconsistency detection benchmarks. First, they manually verify the factual consistency of a small set of seed summaries. Then, they use LLMs to generate numerous edited versions (e.g., via entity modification, antonym swap, hallucinated fact insertion, and negation insertion) of these consistent seed summaries. Finally, human annotators determine whether each edit introduces a factual inconsistency. \citeauthor{laban-salesforce-2023-emnlp-summedits} implement the protocol on ten diverse textual domains, including the legal, dialogue, academic, financial, and sales domains. 

\paragraph{\fib}
\citet{tam-unc-2023-acl-fib} propose a factual inconsistency benchmark, where each example consists of a document and two summaries (one factually consistent summary and one factually inconsistent summary). For factually consistent summaries, they consider reference summaries from \cnndaily~\citep{nallapati-ibm-2016-conll-summarization} and \xsum~\citep{narayan-edinburgh-2018-emnlp-xsum} and manually fix these factually inconsistent reference summaries using minimal edits. They also manually choose factually inconsistent summaries from model-generate summaries.

\end{document}